\pdfoutput=1

\documentclass[11pt]{article}

\usepackage{authblk}
\usepackage{ACL2023}
\usepackage{times}
\usepackage{latexsym}
\usepackage[T1]{fontenc}
\usepackage[utf8]{inputenc}
\usepackage[normalem]{ulem}
\usepackage{microtype}
\usepackage{inconsolata}
\usepackage{graphicx}
\usepackage{caption}
\usepackage{float}
\usepackage{booktabs}
\usepackage{amsmath}
\usepackage{amsfonts}
\usepackage{enumitem}
\usepackage{siunitx}
\usepackage{multirow}
\usepackage{hyperref}
\usepackage{cleveref}
\usepackage{appendix}
\usepackage{subcaption}
\usepackage{csvsimple}
\usepackage{lipsum}
\usepackage{authblk}
\usepackage{pgfplotstable}
\usepackage{filecontents}
\definecolor{LMUGrey3}{HTML}{626468}

\usepackage[framemethod=tikz]{mdframed}

\newmdenv[
  backgroundcolor=LMUGrey3!10,
  linecolor=LMUGrey3!10,
  innertopmargin=2pt,
  innerbottommargin=2pt,
  innerleftmargin=2pt,
  innerrightmargin=2pt,
  skipabove=10pt,
  skipbelow=10pt,
  leftmargin=0pt,
  rightmargin=0pt,
  linewidth=0pt,
]{lightbox}

\title{MultiClimate: Multimodal Stance Detection on Climate Change Videos}

\author{
    \textbf{Jiawen Wang}$^{\diamondsuit,*}$ \enspace
    \textbf{Longfei Zuo}$^{\diamondsuit,*}$ \enspace
    \textbf{Siyao Peng}$^{\diamondsuit,\dagger}$ \enspace
    \textbf{Barbara Plank}$^{\diamondsuit,\dagger}$\\
    $^\diamondsuit$Center for Information and Language Processing, LMU Munich, Germany\\
    $^\dagger$MaiNLP \& MCML, LMU Munich, Germany\\
    {\{jiawen.wang, zuo.longfei\}@campus.lmu.de} \enspace  
    \{siyao.peng, b.plank\}@lmu.de
}

\begin{document}
\maketitle
\def\thefootnote{*}\footnotetext{Equal contributions.}\def\thefootnote{\arabic{footnote}}

\begin{abstract}
  Climate change (CC) has attracted increasing attention in NLP in recent years. However, detecting the stance on CC in multimodal data is understudied and remains challenging due to a lack of reliable datasets. To improve the understanding of public opinions and communication strategies, this paper presents MultiClimate, the first open-source manually-annotated stance detection dataset with $100$ CC-related YouTube videos and $4,209$ frame-transcript pairs. We deploy state-of-the-art vision and language models, as well as multimodal models for MultiClimate stance detection. Results show that text-only BERT significantly outperforms image-only ResNet50 and ViT. Combining both modalities achieves state-of-the-art, $0.747$/$0.749$ in accuracy/F1. Our 100M-sized fusion models also beat CLIP and BLIP, as well as the much larger 9B-sized multimodal IDEFICS and text-only Llama3 and Gemma2, indicating that multimodal stance detection remains challenging for large language models. Our code, dataset, as well as supplementary materials, are available at \url{https://github.com/werywjw/MultiClimate}. 
\end{abstract}

\section{Introduction}

As climate change (CC) gains global attention, measuring human stance towards CC becomes increasingly important.
Numerous large language models (LLMs) and deep learning models have been developed to address these challenges. These models can help detect public opinions and assist stakeholders to improve decision-making, thus providing valuable insights into public perception regarding climate change~\cite{openai2024hello,meta2024introducing,DosovitskiyB0WZ21,AlayracDLMBHLMM22}.

Stance detection is a task to determine whether authors of a document support, oppose, or take a neutral stance toward a specific target~\cite{mohammad-etal-2016-semeval,hardalov-etal-2022-survey,WeinzierlH23}. 
It enhances information management by efficiently categorizing diverse opinions. Identifying varying public opinions helps promote societal understanding and communication, thus reducing conflicts and enhancing public discourse.
Previous research explored stance detection on climate change~\cite{MaynardB15,vaid-etal-2022-towards,DBLP:conf/ijcai/UpadhyayaFN23,DBLP:journals/ipm/UpadhyayaFN23}, but focusing on text-only data. 

\begin{figure}[t]
  \centering
  \begin{subfigure}{\linewidth}
  \includegraphics[width=.47\linewidth]{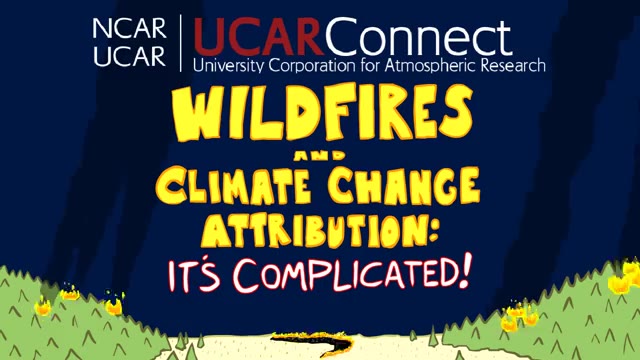}
  \hfill
\includegraphics[width=0.47\linewidth]{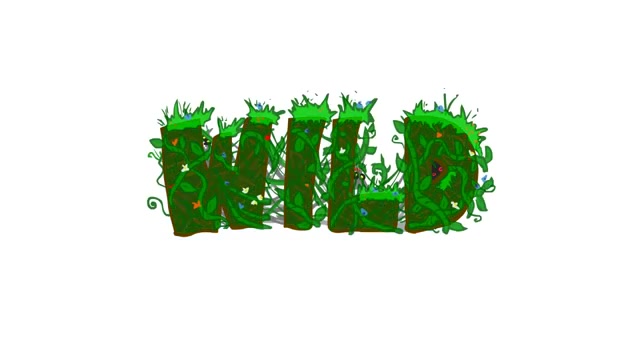}
    \label{subfig:anytime}
  \caption{\uline{Anytime you hear the word} \texttt{[0:05, left, \textsc{Neutral}]} \uline{wild you can bet it's referring to something uncontrollable and unrestrained} \texttt{[0:10, right, \textsc{Oppose}]}.}
  \end{subfigure}
  
  \begin{subfigure}{\linewidth}
\includegraphics[width=.47\linewidth]{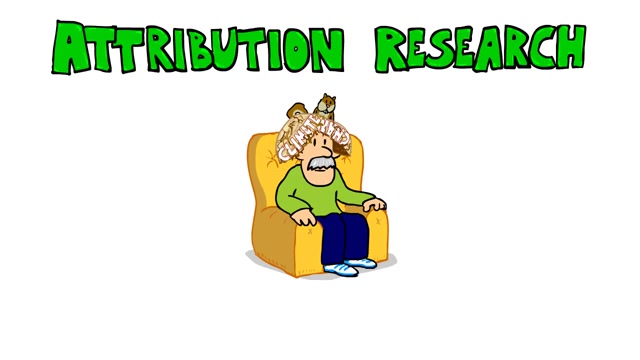}
\hfill
\includegraphics[width=.47\linewidth]{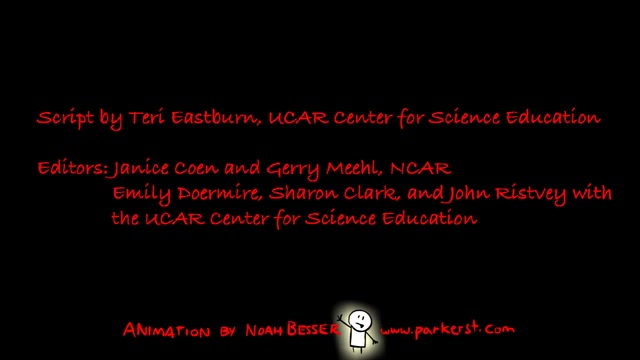}
    \label{subfig:but-sit}
    \caption{But sit tight \uline{attribution research may crack the nut and science researchers in this emerging field are} \texttt{[1:55, left, \textsc{Support}]} \uline{on the case} \texttt{[2:00, right, \textsc{Neutral}]}.}
  \end{subfigure}

  \caption{MultiClimate sample annotations with aligned video frames and transcript sentences.}
  \label{fig:WCCA}
\end{figure}

Images and videos can shape public awareness of climate change by vividly illustrating its social impacts. Recently, multimodality gained significant traction for connecting CC to discourse \citep{WeinzierlH23}, as images and videos largely impact CC  perceptions~\cite{wang2024visual}. 
Investigating stance through multimodality also becomes essential. 
However, due to the lack of datasets, developing multimodal models for stance detection on climate change remains challenging, as previous studies mainly focused on language texts rather than the visual modality. 

To fill this gap, this paper proposes the first open-sourced dataset, MultiClimate, that integrates both visual and textual modalities, specifically $4,209$ image frames and transcripts, to label stances in $100$ CC-related YouTube videos 
(\S\ref{sec:dataset}).
\S\ref{sec:evaluation} evaluates text-only, image-only, and multimodal models on MultiClimate stance detection. 
\S\ref{sec:conclusion} concludes the paper and proposes future directions.

Our results 
show that the text-only BERT model outperforms image-only models on multimodal stance detection, and the best performance is achieved by fusing models from both modalities. 
We further experiment with 9B-sized large language and multimodal models and illustrate that these larger models deliver unsatisfactory zero-shot results, much lower than our state-of-the-art (SOTA) fusion models.
Fine-tuning a large multimodal model brings about some improvements, but it is resource-heavy and encounters the Green NLP problems. 

\section{Related Work}
\label{sec:related}

\paragraph{Multimodal Stance Detection.}
Stance detection has mainly concentrated on textual analysis~\cite{KucukC20, LanGJ024}, with a significant focus on the stance expressed in social media platforms like Twitter~\cite{TaulePMR18,ConfortiBPGTC20}. 
Yet, a recent trend arose that gradually includes images and videos in stance detection~\cite{KucukC21, CarnotHBSKFP023}. 

Current multimodal stance detection datasets emphasize different aspects such as communication frames and trending topics like COVID-19 \citep{TaulePMR18,weinzierl-harabagiu-2023-identification,liang-etal-2024-multi}. 
These data rely solely on static images or extract the first frame from a video or GIF for visual input.
Despite these advancements, research on stance detection using visual input remains limited.
Our MultiClimate dataset addresses climate change topics by utilizing frames from full videos as visual input and transcripts as text, allowing for a more comprehensive information coverage.

\paragraph{Stance Detection on Climate Change.} 
Stance detection on CC~\cite{fraile-hernandez-penas-2024-hamison} aims at determining whether a given document expresses a supportive, opposing, or neutral attitude toward whether CC is a real concern. 
Recent CC stance detection studies particularly focused on social media texts~\cite{vaid-etal-2022-towards}, especially Twitter~\cite{ConfortiBPGTC20,DBLP:conf/ijcai/UpadhyayaFN23}. 

Regarding modeling, for instance, \citet{vaid-etal-2022-towards} offered Fast-Text~\cite{bojanowski-etal-2017-enriching} and BERT~\cite{devlin-etal-2019-bert} variants for stance tasks. 
\citet{DBLP:conf/ijcai/UpadhyayaFN23} proposed a framework that utilizes emotion recognition and intensity prediction to discern different attitudes in tweets about climate change. CC stance detection is also applied to downstream tasks such as fake news detection~\cite{MazidZ22}.
However, none of the models above handles multimodal inputs. 

\section{The MultiClimate Dataset}
\label{sec:dataset}

We propose a new open-source MultiClimate dataset, the first climate change stance detection corpus on multimodal data.
To the best of our knowledge, Mendeley provides a closest dataset to ours~\cite{aharonsondata}.
However, it only provides a simple Excel file with $168$ YouTube links and some basic statistics on each video.\footnote{\url{https://data.mendeley.com/datasets/j955mxnyyf/1}}
Most of these videos are not under the Creative Commons license and none is annotated with stance labels.

\begin{table}[t]
  \centering
\resizebox{\columnwidth}{!}{ 
  \begin{tabular}{l|r|rrr|r}
    \toprule
    & \multirow{2}{*}{\textbf{Videos}} & \multicolumn{4}{c}{\textbf{Frame-Transcript Pairs}} \\
   &  & \textsc{Support} & \textsc{Neutral}& \textsc{Oppose} & \textbf{Total} \\
    \midrule
    \textbf{Train} &$80$  & $1,449$ & $1,036$& $887$  & $3,372$ \\
    \textbf{Dev} & $10$  & $204$& $83$  & $130$  & $417$ \\
    \textbf{Test} & $10$   & $194$ & $73$& $153$ & $420$ \\
    \midrule
    \textbf{Total} & $100$  & $1,847$ & $1,192$ & $1,170$ & $4,209$ \\
    \bottomrule
  \end{tabular}
  }
  \caption{Overview of MultiClimate 
  statistics.}
  \label{tab:statistics}
\end{table}

Our MultiClimate dataset consists of $100$ CC-related YouTube videos in English with $4,209$ frame-transcript pairs.
To ensure a balanced distribution, we partition MultiClimate into $80\%$ train, $10\%$ dev, and $10\%$ test sets, not only in terms of the number of videos but also the number of frame-transcript pairs.
\Cref{tab:statistics} provides statistics on MultiClimate frame-transcript pairs. 
For a complete list of $100$ YouTube videos, see \Cref{app:dataset}.

\paragraph{Video Collection.}
100 YouTube videos are collected by querying  ``climate change'' and filtering the results by Creative Commons license. The videos are downloaded using the \texttt{pytube}\footnote{\url{https://pytube.io/en/latest/}} and \texttt{yt-dlp}\footnote{\url{https://github.com/yt-dlp/yt-dlp}} APIs, while the transcripts are extracted using the \texttt{youtube-transcript-api}.\footnote{\url{https://pypi.org/project/youtube-transcript-api/}}

\paragraph{Frame Extraction and Alignment.}
To effectively use our video data, we extract the initial frame from each $5$-second segment using \texttt{FFmpeg} \cite{tomar2006converting}. 
We then manually align each frame to its corresponding full sentence in the transcripts to form a frame-transcript pair.
Longer sentences can be aligned with 
multiple time-stamped frames,\footnote{
MultiClimate has $1,902$ transcripts aligned to 1 frame, $656$ to 2 frames, 
$193$ to 3, 
$27$ to 4, 
$8$ to 5, 
$3$ to 6, 
$2$ to 7, 
excluding \texttt{[Music]} or \texttt{None} transcripts. Details are provided in the repository \url{https://github.com/werywjw/MultiClimate/tree/main/notebooks}.} 
and we allow different stance labels for individual frame-transcript pairs, as shown in \Cref{fig:WCCA}.

\paragraph{Stance Annotation.}
The first two authors of this paper manually annotated the stance expressed in each frame-transcript pair for $100$ videos.
We use trinary stance labels: \textsc{Support}, \textsc{Neutral}, and \textsc{Oppose}.
\Cref{app:annotation} details annotation guidelines, particularly label definitions, and examples. 

MultiClimate includes $1,847$ \textsc{Support}, $1,170$ \textsc{Oppose}, and $1,192$ \textsc{Neutral} frame-transcript pairs (\Cref{tab:statistics}).
To assess the effectiveness of our annotation guideline and the quality of our dataset, all $10$ videos in the test partition are double annotated. 
Inter-annotator agreement (IAA) between the two annotators achieves $0.703$ in Cohen's kappa, $0.826$ in accuracy, and $0.823$ in weighted F1; see \Cref{app:iaa} for IAA on $10$ individual test videos.
Additionally, we aggregate stance labels from each video's frame-transcript pairs by majority voting for future video-level analyses and experiments; 
\Cref{app:dataset} includes these video stance labels.

\section{MultiClimate Stance Detection}
\label{sec:evaluation}

This section evaluates SOTA text-only, image-only, and multimodal models 
on MultiClimate stance detection. 
We use both accuracy and weighted F1 scores since \textsc{Support/Neutral/Oppose} labels are unbalanced in the dataset. The models are run on the CPU, Google T4 GPU, and NVIDIA GeForce RTX 2080.

\subsection{Models}
\label{subsec:models}

We conduct a comprehensive performance evaluation on several text-only, image-only, text-image-fusion, and multimodal models. 
We leverage text-only BERT variants~\citep{devlin-etal-2019-bert} as earlier work showed their superior performance on tweet stance detection~\cite{weinzierl-harabagiu-2023-identification,vaid-etal-2022-towards}. 
Large language models (LLMs) are also included given their promising zero-shot classification performance~\cite{dubey2024llama}. 

Meanwhile, for image recognition and analysis, ResNet50~\citep{HeZRS16} and ViT~\citep{DosovitskiyB0WZ21} are capable of understanding and interpreting complex image data. 
By combining aforementioned textual and visual models, we also deploy fusion models to investigate whether both modalities are essential for multimodal stance detection. 
Moreover, models that are trained on cross-modal representations, CLIP~\citep{RadfordKHRGASAM21},BLIP~\citep{Li2022BLIP}, and IDEFICS~\citep{AlayracDLMBHLMM22}, are also compared with the aforementioned unimodal and fusion models.

\paragraph{Text-only Models.}

We use BERT (Bidirectional Encoder Representations from Transformers, \textit{bert-base-cased}, \citealt{devlin-etal-2019-bert}) for our textual stance detection given its effectiveness. Newly released LLMs, Llama3 (\textit{meta-llama/Meta-Llama-3-8B}, \citealt{dubey2024llama,meta2024introducing}) and Gemma2-9B (\textit{google/gemma-2-9b}, \citealt{google2024gemma2}) are also evaluated 
on the Ollama~\cite{ollma2024} platform
by giving the following zero-shot prompt: 
\begin{lightbox}
  \textit{\small Classify the stance of the following text towards climate change as: 0 (\textsc{Neutral}), 1 (\textsc{Support}), 2 (\textsc{Oppose}): \{transcript\}}.
\end{lightbox}

\paragraph{Image-only Models.}
We also deploy two state-of-the-art image-only models, ResNet50 (Residual Network, \textit{microsoft/resnet-50}, \citealt{HeZRS16}) and
ViT (Vision Transformer, \textit{google/vit-base-patch16-224}, \citealt{DosovitskiyB0WZ21}).

\paragraph{Multimodal Models.}

Our multimodal fusion models are built by concatenating BERT~\cite{devlin-etal-2019-bert} with ViT~\citep{DosovitskiyB0WZ21} or ResNet50~\cite{HeZRS16} embeddings, as they are the smaller 100M-sized models (see \Cref{tab:results}). 

CLIP (Contrastive Language-Image Pre-training, \textit{openai/clip-vit-base-patch32}, \citealt{RadfordKHRGASAM21}) and BLIP (Bootstrapping Language-Image Pre-training, \textit{Salesforce/blip-image-captioning-base}, \citealt{Li2022BLIP}) are leveraged to associate images and text simultaneously, capturing richer, more nuanced information. 
We also experiment with 
IDEFICS (Image-aware Decoder Enhanced à la Flamingo with Interleaved Cross-attentionS,
\textit{HuggingFaceM4/idefics-9b}, \citealt{AlayracDLMBHLMM22}),\footnote{\url{https://huggingface.co/blog/idefics}} an open-source Multimodal Large Language Model (MLLM) by providing the following prompt template in zero-shot as well as fine-tuned settings.

\begin{lightbox}
  \textit{\small 
   Given the \{frame\} and \{transcripts\}, what is the stance of this frame-transcript pair towards climate change? Choose one between 0 for \textsc{Neutral}, 1 for \textsc{Support}, and 2 for \textsc{Oppose}.
   }
\end{lightbox}

\subsection{Results and Discussions}
\Cref{tab:results} presents the results of evaluating the effectiveness of individual modalities, multimodal models, and 9B-sized large models on MultiClimate.

\begin{table}[t]
  \centering
  \resizebox{\columnwidth}{!}{
    \begin{tabular}{r|cc|r}
        \toprule
        \textbf{Model}& \textsc{Acc} & \textsc{F1} & \# Params \\
        \midrule
        \textbf{BERT}$^{\clubsuit}$   & $\textbf{0.705}$ & $\textbf{0.705}$ & 110M\\
        \textbf{Llama3}$^{\clubsuit}$ (zero-shot) & $0.485$ & $0.451$ & 8B\\
        \textbf{Gemma2}$^{\clubsuit}$ (zero-shot)  & $0.461$ & $0.382$ & 9B\\
        \midrule
        \textbf{ResNet50}$^{\spadesuit}$ & $0.424$ & $0.399$  & 25.6M\\
        \textbf{ViT}$^{\spadesuit}$ & $\textbf{0.460}$ & $\textbf{0.462}$ & 86.6M \\ 
        \midrule
        \textbf{BERT + ResNet50}$^{\star}$ & $0.717$ & $0.714$ & 111.7M\\
        \textbf{BERT + ViT}$^{\star}$ & $\textbf{0.747}$ & $\textbf{0.749}$ & 196.8M\\
        \textbf{CLIP}$^{\star}$ & $0.431$ & $0.298$ &151.3M\\
        \textbf{BLIP}$^{\star}$ & $0.462$ & $0.292$ & 470M \\
        \textbf{IDEFICS}$^{\star}$ (zero-shot) & $0.347$ & $0.270$ & 9B\\
        \textbf{IDEFICS}$^{\star}$ (fine-tuned) & $0.600$ & $0.591$ & 9B\\
        \midrule
        \textsc{Human} &   $\textbf{0.826}$ & $\textbf{0.823}$ & - \\ 
        \bottomrule
    \end{tabular}
    }
    \caption{Text-only$^{\clubsuit}$, image-only$^{\spadesuit}$, and multimodal$^{\star}$ model results on the MultiClimate test set.}
  \label{tab:results}
\end{table}

\paragraph{BERT results are outstanding.}
The text-only BERT model achieves the best performance among single-modal models, notably surpassing the zero-shot LLMs. 
Furthermore, the multimodal fusion model BERT + ViT achieves state-of-the-art, $0.747$ in accuracy and $0.749$ in F1 score. Generally speaking, BERT + ResNet50/ViT fusion models outperform CLIP, BLIP, and IDEFICS, as textual features are crucial to our CC stance detection, and transcripts in YouTube videos benefit the already well-performing BERT model. 

CLIP performs the worst in accuracy among trained multimodal models. 
One hypothesis is that the maximum sequence length has an impact on the results, 
that is, the maximum sequence length of CLIP is $77$ tokens for text inputs, much shorter than BERT, and leads to declined performance due to a lack of information. 
For instance, the ``MACC'' video includes one sentence with $82$ tokens, exceeding CLIP's limit (77 tokens) but not BERT's.
Notably, fine-tuned BLIP shows similar poor performance, in particular low weighted F1 score, which can be attributed to the misclassification of minority classes such as \textsc{Neutral} and \textsc{Oppose}.

\paragraph{Textual and visual information compensate each other.} 
\Cref{tab:results} shows text-only models overall perform considerably better than image-only models, 
indicating that specific language contextual understanding can outperform approaches that are trained for detailed image classification. 
However, if both visual and textual information are concatenated, we can achieve the optimal result. 
We note that even though transcripts generally contain richer linguistic information, it is not guaranteed that visual and textual information are both meaningful in every frame-transcript instance. 
For example, video segments with no speech but only music playing in the background are transcribed as \texttt{[Music]} in YouTube. 
Text-dependent models can barely capture stances from text data in such cases, while the image can reveal additional information, and hence visual-informed models predict labels that align with human annotations; see the last $6$ frame-transcript pairs of ``AMCC'' in \Cref{app:llm}. 

\paragraph{Text-only Llama3 wins in zero-shot.}
We also observe that Llama3 performs better than single-modal Gemma2 and multimodal IDEFICS within zero-shot, 
with a marginally increase of $0.138$ in accuracy and $0.181$ in F1 score compared with 9B-sized IDEFICS. 
Since Llama3 has been trained on more extensive text and speech data from various domains, 
it can better process longer sequences of text~\cite{dubey2024llama}. 
Meanwhile, we hypothesize that Llama3 is better at handling noisy data, which is common in the transcripts as many videos feature colloquial speech (e.g., in ``ACCFP''), thereby increasing the noise level. 
The results, alike the earlier BERT superior performances, suggest that the textual part is vital in CC stance detection due to its more explicit narrative and clearer directionality, and leveraging SOTA LLMs can significantly improve performance. 

\paragraph{Zero-shot IDEFICS is biased toward \textsc{Support}.} We also observe a tendency for the zero-shot IDEFICS model to predominantly predict \textsc{Support} labels, less often \textsc{Neutral}, and rarely \textsc{Oppose}. 
This bias impacts the model's performance negatively on videos where the majority of gold labels are \textsc{Oppose} or \textsc{Neutral}. 
This accounts for the poor performance observed in videos like ``CCUIM'' (Acc/F1: $0.167/0.111$; $7$ \textsc{Support}, $21$ \textsc{Neutral} and 20 \textsc{Oppose}) and ``EWCC'' (Acc/F1: $0.163/0.109$; $9$ \textsc{Support}, $11$ \textsc{Neutral} and $29$ \textsc{Oppose}).
In contrast, the human inter-annotator Acc/F1 scores on these two test documents are not low: $0.771$/$0.773$ on ``CCUIM'' and $0.816$/$0.811$ on ``EWCC,'' illustrating that such bias is not present during human annotation.

The underlying reason for this annotation bias can stem from the mix of ``stance'' and ``sentiment''. 
During manual annotation, instances that are negatively framed are often categorized as \textsc{Oppose}. 
However, the model classifies based on the expressed ``stance'' towards CC. 
Given that our video selection was filtered under ``climate change'', a \textsc{Support} stance predominates. 
Additionally, the model occasionally conflates frame information, 
whereas annotations are based on both the frame and paired transcripts. Consequently, frames opposing climate change are undervalued in the classification of the model, 
leading to a pronounced bias towards \textsc{Neutral} and \textsc{Support} categories.

\paragraph{Fine-tuned IDEFICS reduces stance bias.}

To mitigate biases present in the zero-shot IDEFICS model, 
we fine-tune IDEFICS using LoRA (Low-Rank Adaptation, \citealt{HuSWALWWC22}) 
on 80/10 MultiClimate train/dev videos before evaluation;
see \Cref{app:hyper} for a list of hyperparameters.

Fine-tuning results in a significant increase in model performance. The accuracy increases from $0.347$ to $0.600$, and 
the F1 score improves from $0.270$ to $0.591$. These enhancements suggest that the model has developed a stronger capability for CC stance detection.
Besides the higher Acc/F1, after examining the predictions made by the model before and after the fine-tuning process (see \Cref{app:cm}), 
we observe a marked increase in the number of instances classified as \textsc{Oppose}, particularly the percentage of correctly predicted gold \textsc{Oppose} labels increased from $0.00\%$ to $83.01\%$, indicating a better performance and less model bias.

\section{Conclusion and Future Directions}
\label{sec:conclusion}

This paper curates MultiClimate, a novel multimodal stance detection dataset to support video research on climate change. 
Our results show that BERT-fusion models can achieve considerably higher performance than large multimodal models. Textual information is vital on MultiClimate, 
while the visual modality can compensate for the drawbacks of language models, and the best is achieved when combining image and text information. 
Similarly, SOTA LLMs beat large multimodal models in zero-shot. 
Our work provides a foundation for multimodal stance detection in CC. 

We plan to extend our dataset with more videos and annotations 
and further investigate the interactions between visual and textual features, as well as the informativeness of each modality in manual stance labeling. 
We are also interested in expanding to audio and video modalities, as well as unimodal models using transfer learning techniques, to improve performance and explain stance detection in CC. We hope our work fosters the positive social impact of  CC stance detection in the NLP field. 

\section*{Limitations}
While our study makes valuable contributions by enriching the multimodal climate change dataset and exploring the performance of different modalities in stance detection tasks, it is important to recognize several inherent limitations.
First, the dataset annotation was conducted by a limited number of annotators, which may introduce personal biases into the data. 
Second, large language models are not fine-tuned due to limited computational resources.
Third, our study focuses solely on transcripts and frames. Incorporating audio and video modalities can enhance the understanding of speaker's emotions and intentions, and potentially further improve stance detection.

\section*{Acknowledgement}
We thank anonymous reviewers for their constructive feedback on this work. 
This research project is in parts funded by Deutschlandstipendium, AWS (Amazon Web Service).
This work also partially belongs to the KLIMA-MEMES project funded by the Bavarian Research Institute for Digital Transformation (bidt), an institute of the Bavarian Academy of Sciences and Humanities.
The authors are responsible for the content of this publication.

\bibliography{anthology}
\bibliographystyle{acl_natbib}

\newpage
\appendix

\section{Annotation Guideline}
\label{app:annotation}
Annotators are tasked with determining the stance on climate change within each frame-transcript pair provided. This involves assessing both the visual and textual elements of each pair and following the below instructions.

\paragraph{Annotation Order.} Annotators annotate at the frame-transcript level for each pair within a single video, deliberately avoiding sentence context. 

\paragraph{Text and Frame Consideration.} Annotators evaluate both the text and the accompanying image frame. In cases of conflict between text and image, prioritize the element that evokes stronger emotions related to the stance.

\begin{figure*}[h]
    \centering
    \begin{minipage}[t]{\columnwidth}
        \centering
        \includegraphics[width=.95\textwidth]{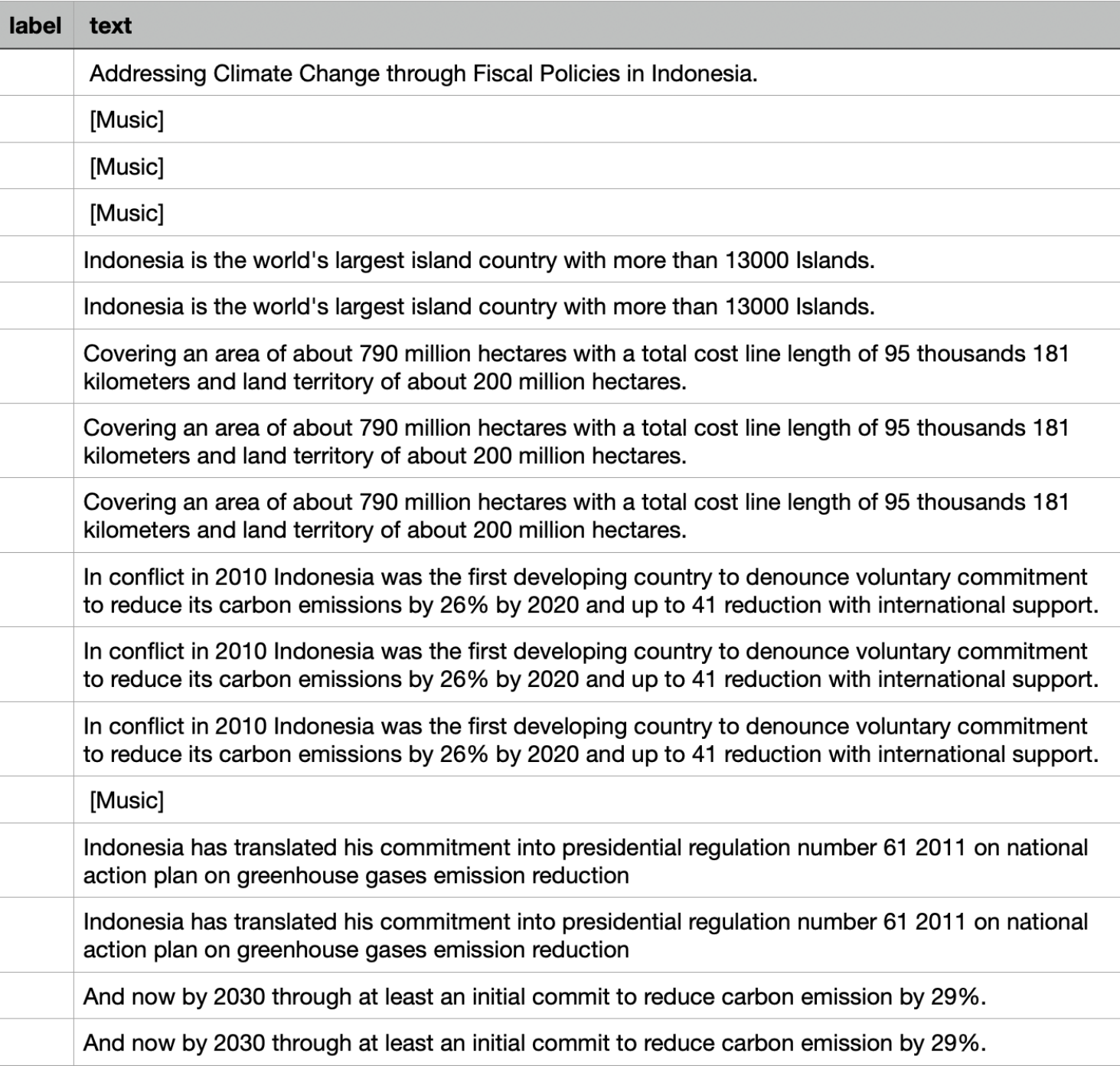}
    \end{minipage}
    \hfill
    \begin{minipage}[t]{\columnwidth}
        \centering
        \includegraphics[width=.95\textwidth]{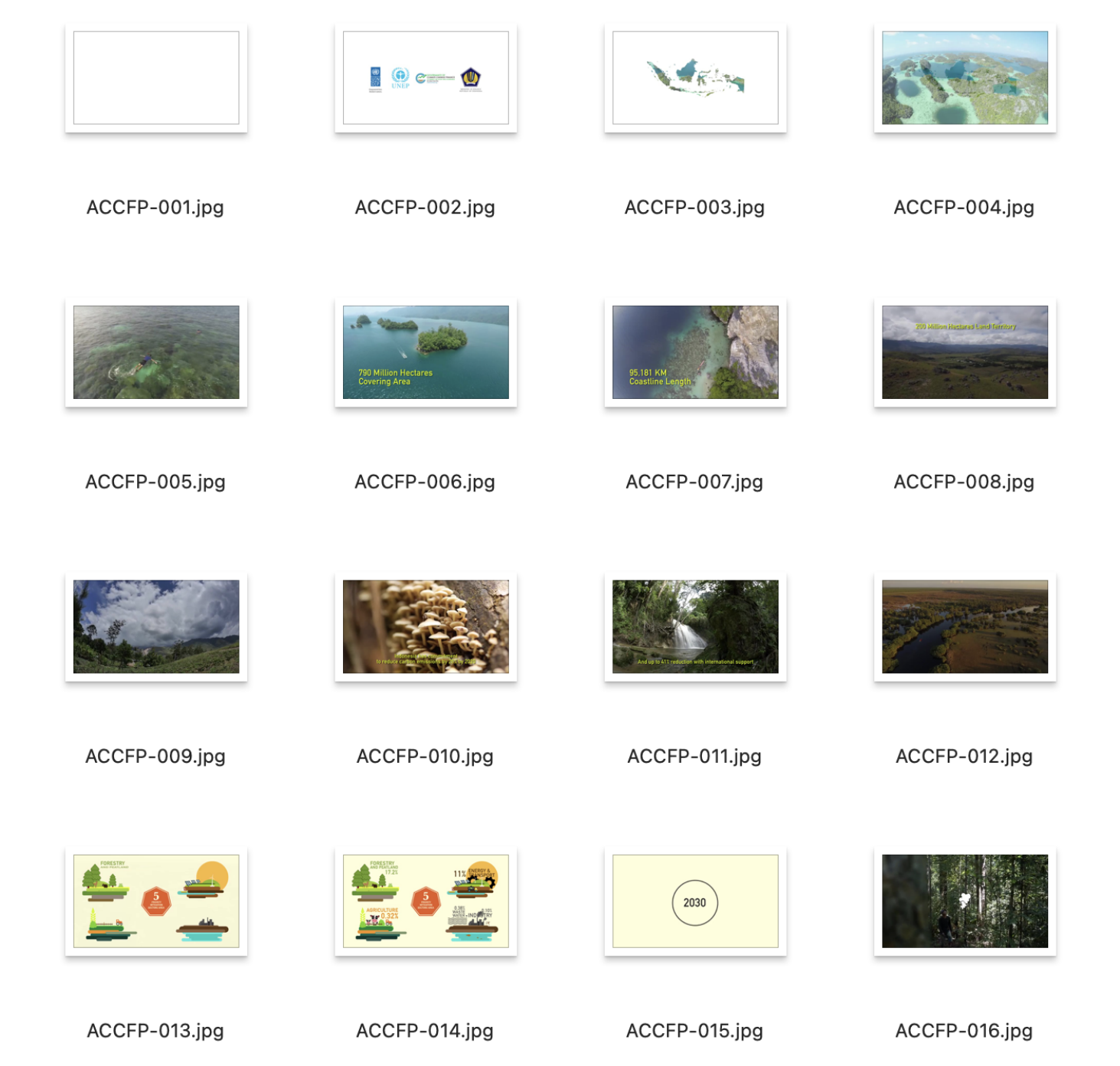}
    \end{minipage}
    \caption{Annotator materials in ``ACCFP'': CSV file (left) and frames (right) provided.}
    \label{fig:annotation}
\end{figure*}

\paragraph{Data Storage.} The annotated dataset is provided in the following formats on the GitHub repository.\footnote{\url{https://github.com/werywjw/MultiClimate/tree/main/dataset}}
As shown in \Cref{fig:annotation}, each sentence transcript is presented in a \texttt{CSV} file with the column label `text', and the corresponding frame is provided as a \texttt{JPEG} file. The sentences and frames are in the same order.

\subsection{Stance Definitions and Examples}
Each frame-transcript pair must be annotated with one of the following stance values regarding climate change: \textsc{Support}, \textsc{Neutral}, or \textsc{Oppose}.

\paragraph{Support.} The frame-transcript pair accepts, agrees with, and/or promotes climate change-related topics or actions. Presented below are several sample categories that exemplify support arguments. Additional categories may also exist.

\begin{itemize}
\setlength\itemsep{0pt}
\item  \textit{Action Promotion}: ``Goal 13 of the sustainable development goals climate action.'' {\small \texttt{WISE-002}}
\begin{center}
\includegraphics[width=0.9\linewidth]{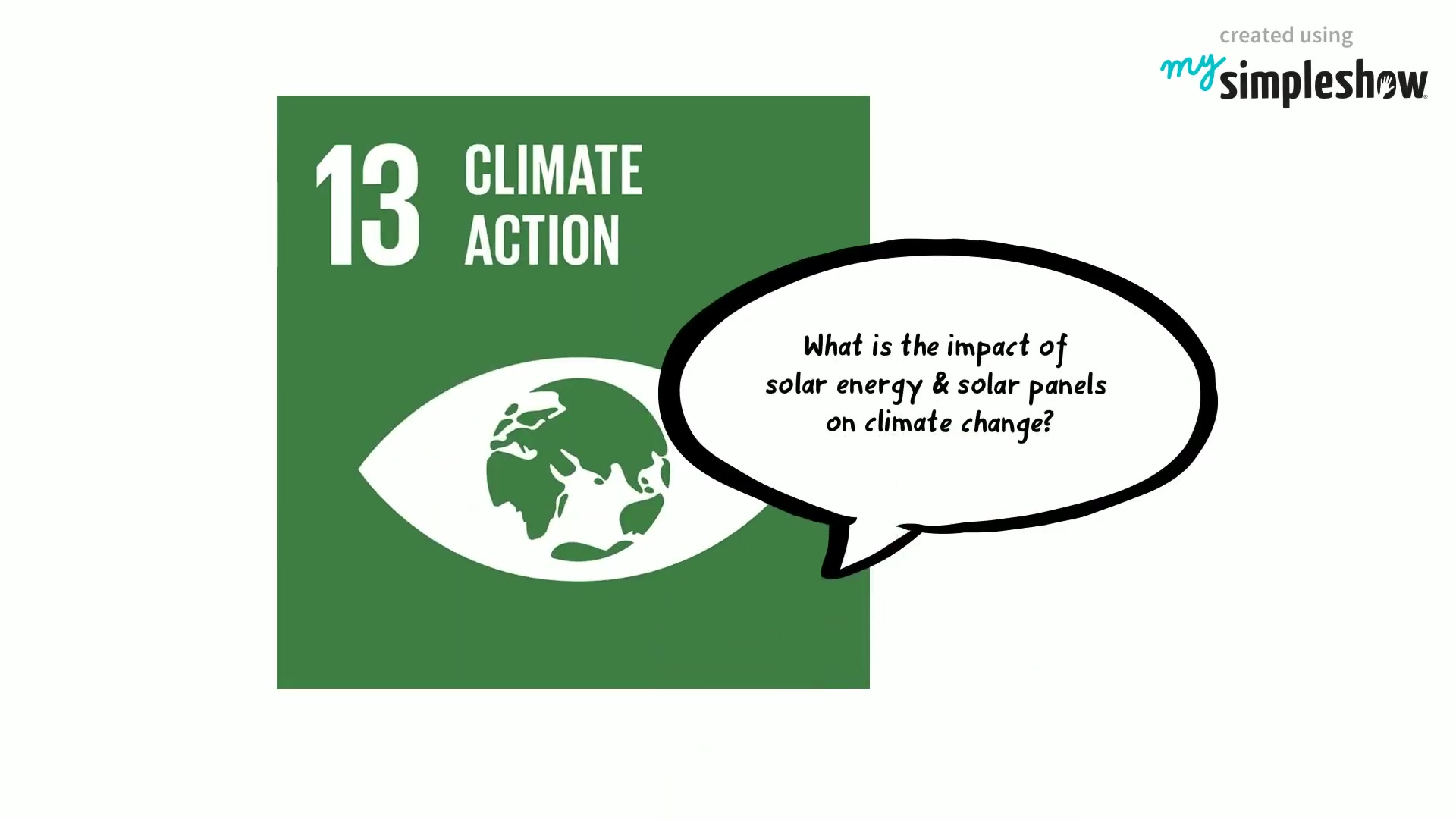}
\end{center}

\item \textit{Encouragement}: ``It's up to us to preserve these natural wonders and maintain the balance on earth.''
{\small \texttt{HCCAE-029}}
\begin{center}
\includegraphics[width=0.9\linewidth]{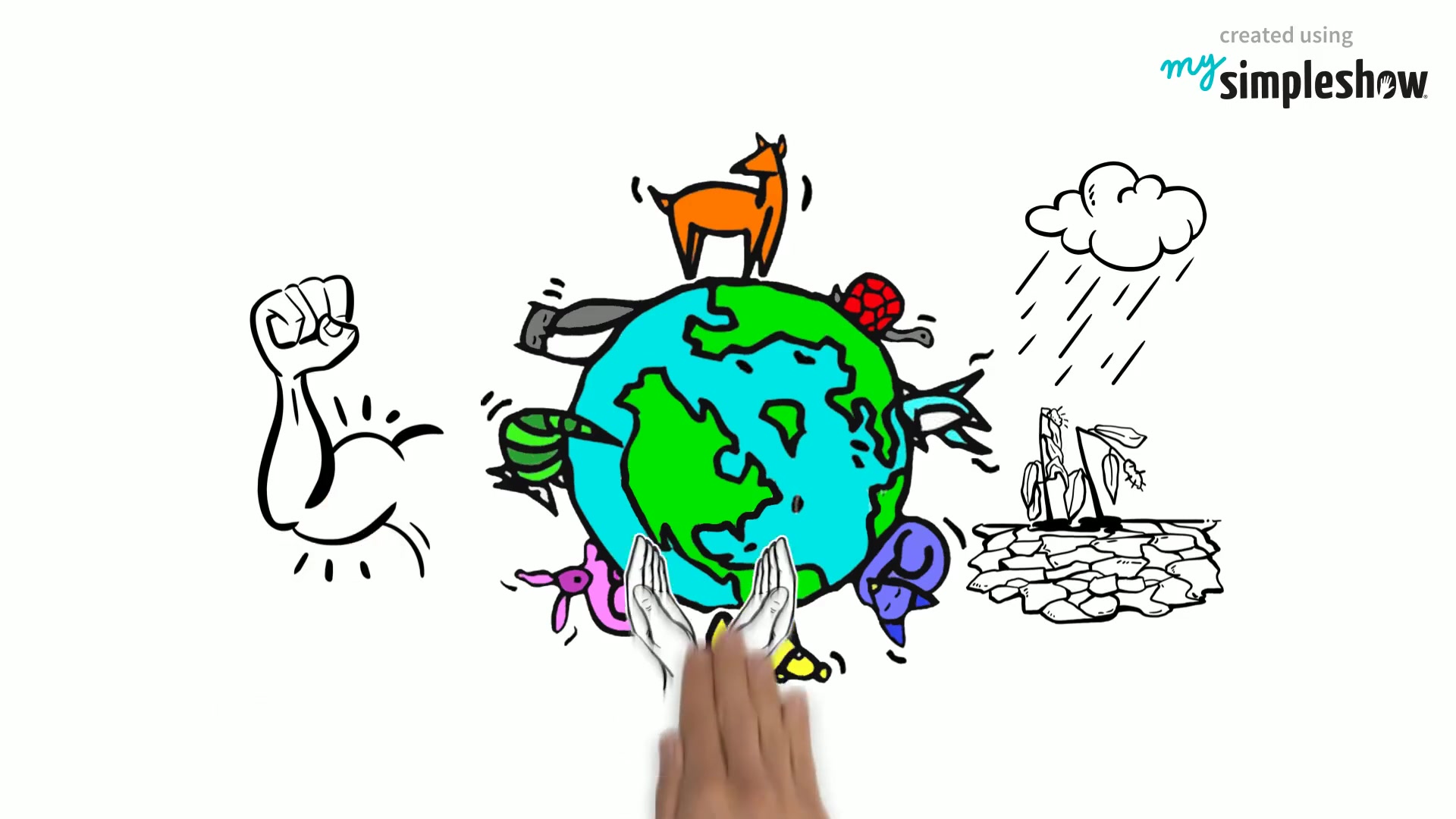}
\end{center}

\item 
\textit{Achievements}: ``It could prevent as much as 174 million tons of carbon from getting released.''
{\small \texttt{DACC-019}}
\begin{center}
\includegraphics[width=0.9\linewidth]{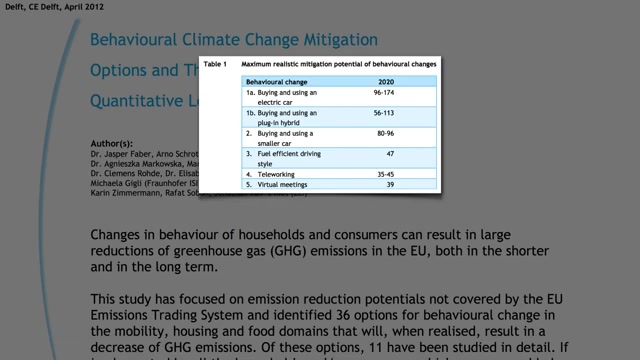}
\end{center}

\item \textit{Solutions}: ``By eating more vegetables and less meat you not only get to enjoy the wonderful variety of fresh produce, but you also help the planet.''
{\small \texttt{CCTA-010}}
\begin{center}
\includegraphics[width=0.9\linewidth]{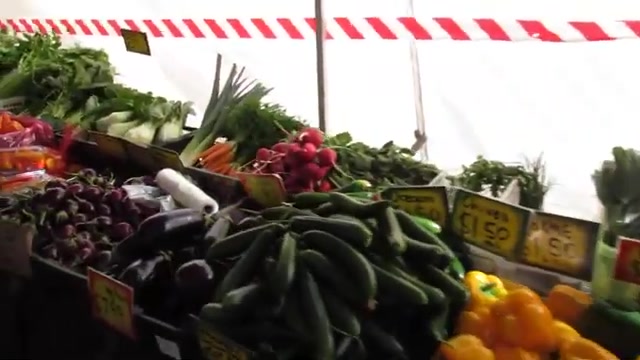}
\end{center}
\end{itemize}

\paragraph{Neutral.} The pair neither supports nor opposes climate change topics or related actions. Presented below are several categories that exemplify neutral arguments. Additional categories may also exist.
\begin{itemize}
\setlength\itemsep{0pt}

\item 
\textit{Unrelated Context}: ``People think that economists don't agree about anything.''
{\small \texttt{RHTCC-025}}
\begin{center}
\includegraphics[width=0.9\linewidth]{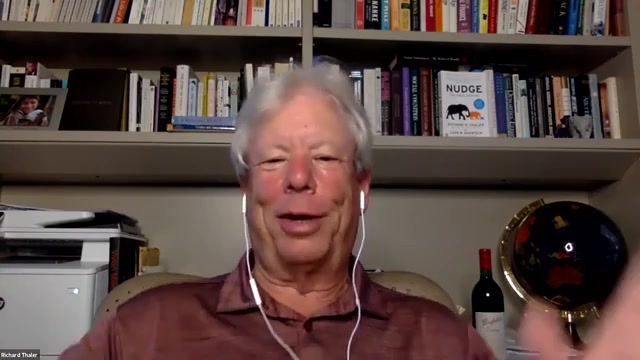}
\end{center}

\item 
\textit{General Information}: ``An ice core is a continuous section of ice drilled into a glacier or an ice sheet.''
{\small \texttt{SDDA-007}}
\begin{center}
\includegraphics[width=0.9\linewidth]{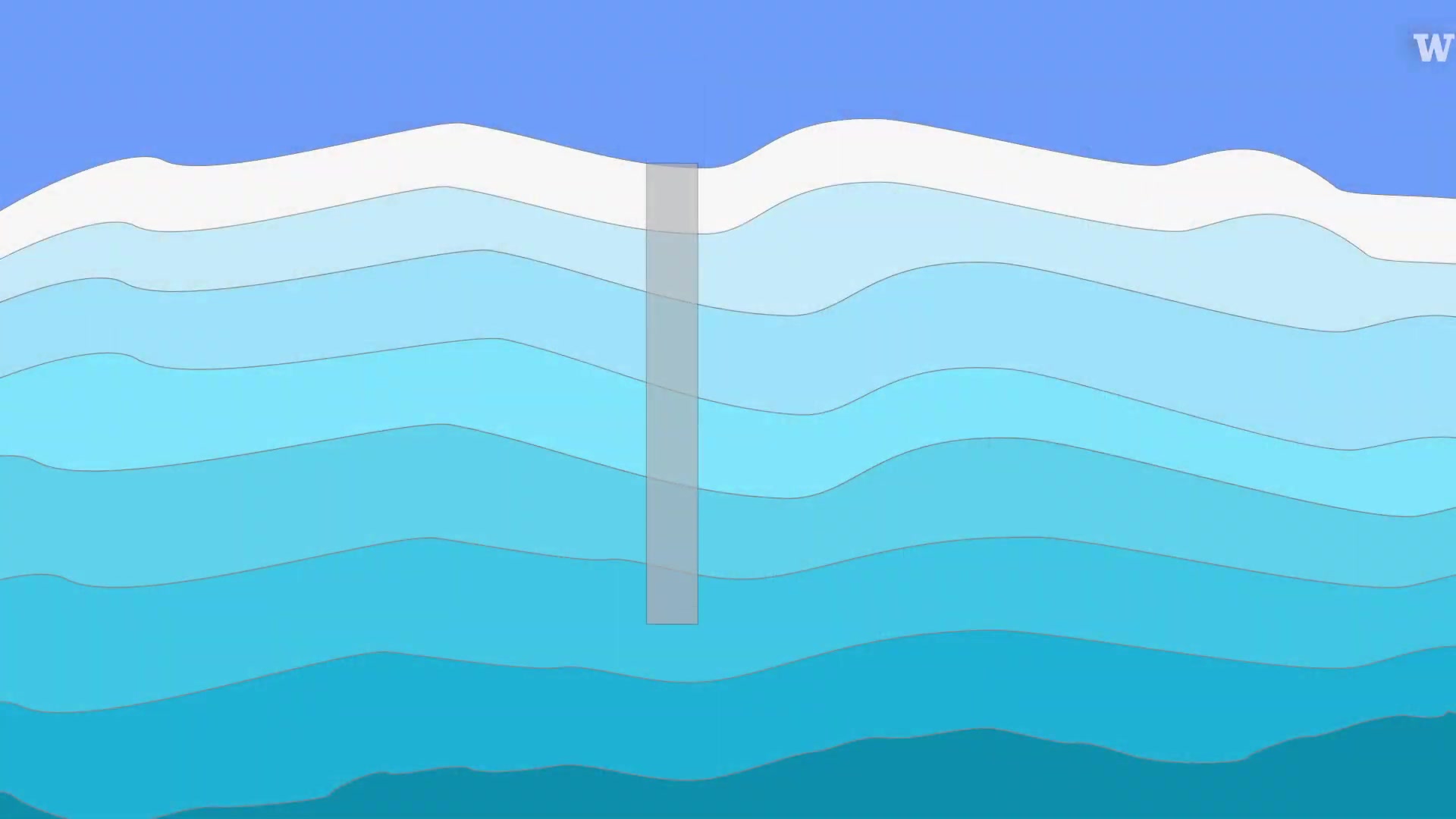}
\end{center}

\item  
\textit{Interrogative}: ``Why have so few thought leaders made it their signature issue?''
{\small \texttt{CCIS-008}}
\begin{center}
\includegraphics[width=0.9\linewidth]{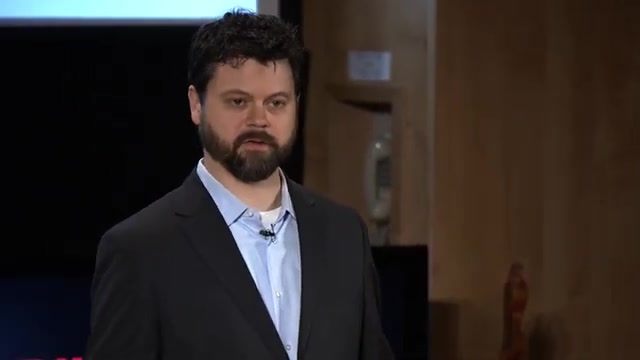}
\end{center}

\end{itemize}

\paragraph{Oppose.} The pair expresses negative sentiments or criticism towards climate change or its related aspects. Presented below are several sample categories that exemplify opposing arguments. Additional categories may also exist.
\begin{itemize}
\item 
\textit{Negative Consequences}: ``Either everyone is leaving, or everyone is killing each other it all comes down to whether there’s enough rain.''
{\small \texttt{TIOCC-005}}
\begin{center}
\includegraphics[width=0.9\linewidth]{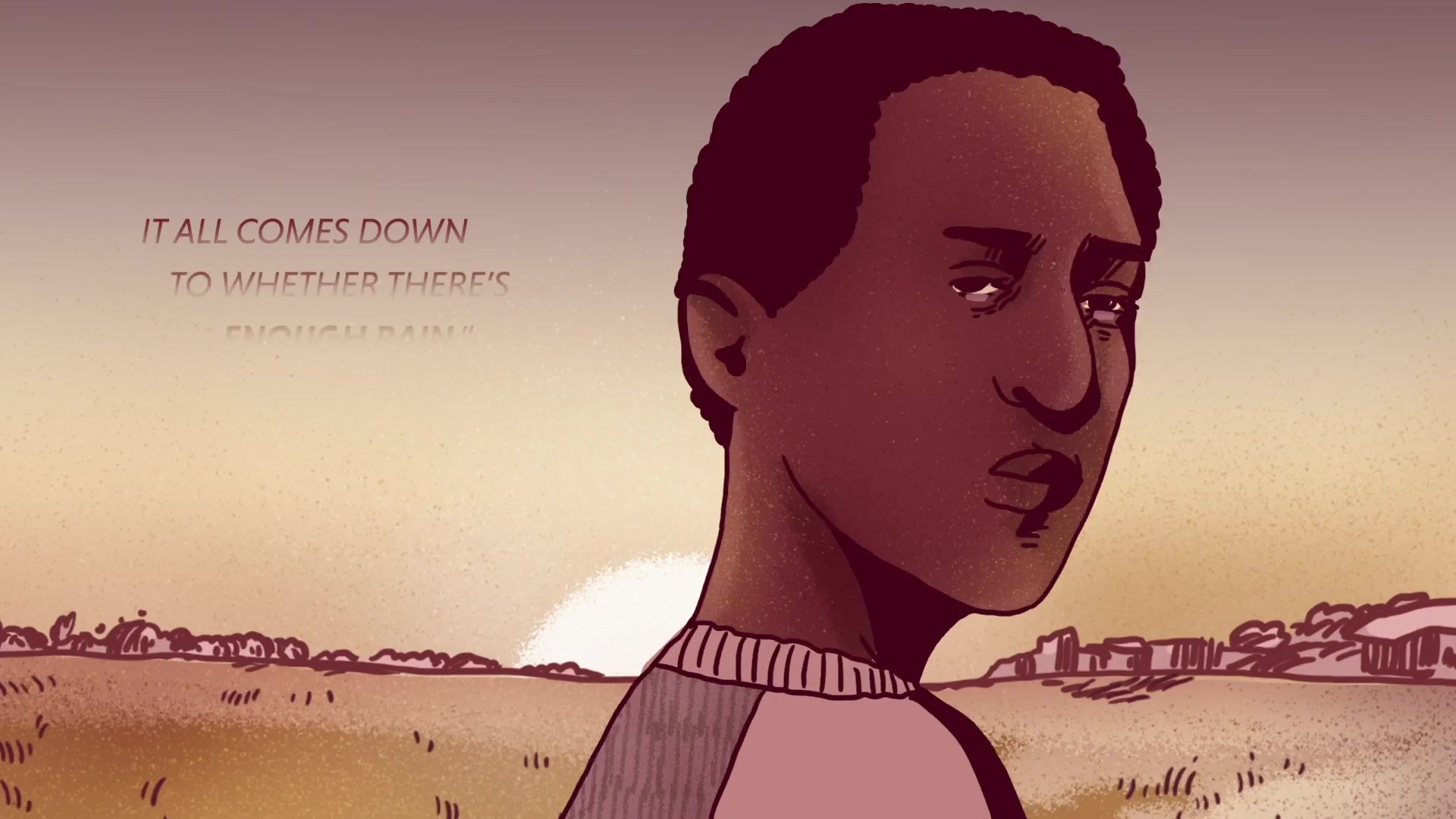}
\end{center}

\item 
\textit{Adverse Effects}:  ``And this is the negative effects the diet is having on the ecosystem and the adverse effects on human health.'' 
{\small \texttt{DACC-047}}
\begin{center}
\includegraphics[width=0.9\linewidth]{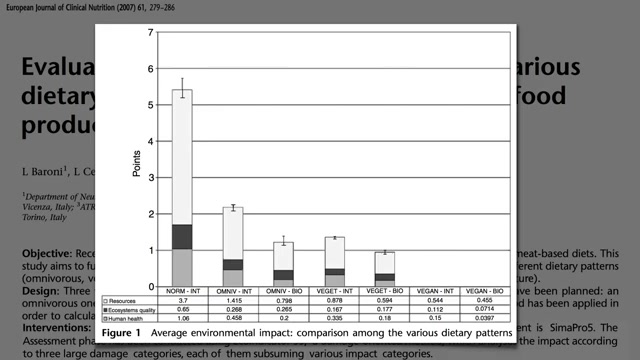}
\end{center}

\end{itemize}

\section{IAA Details on 10 Test Videos}
\label{app:iaa}
We also assessed the accuracy and weighted F1 score on the $10$ test video documents in \Cref{tab:iaa}.

\begin{table}[ht]
  \centering
  \resizebox{0.75\columnwidth}{!}{ 
  \begin{tabular}{@{}cccc@{}}
    \toprule
    \textsc{Video} & \textsc{Cohen's} $\kappa$ & \textsc{Acc} & \textsc{F1}\\
    \midrule
    ACCFP & $0.698$ & $0.867$ & $0.873$ \\
    CCAH & $0.778$ & $0.867$ & $0.851$\\
    CCSAD & $0.408$ & $0.644$ & $0.675$\\
    CCUIM & $0.633$ & $0.771$ & $0.773$\\
    EIB & $0.647$ & $0.822$ & $0.809$\\
    EWCC & $0.690$ & $0.816$ & $0.811$\\
    GGCC & $0.736$ & $0.827$ & $0.814$\\
    SCCC & $0.724$ & $0.824$ & $0.806$\\
    TICC & $0.872$ & $0.936$ & $0.929$\\
    WICC & $0.838$ & $0.900$ & $0.890$\\ 
    \bottomrule
  \end{tabular}
  }
  \caption{Overview of IAA (Cohen's kappa), accuracy, and weighted F1 score on 10 MultiClimate test videos.}
  \label{tab:iaa}
\end{table}

\section{IDEFICS Confusion Matrix of Predictions Before and After Fine-tuning}
\label{app:cm}

\begin{figure}[H]
  \centering
  \resizebox{.95\columnwidth}{!}{   
  \includegraphics[width=\linewidth]{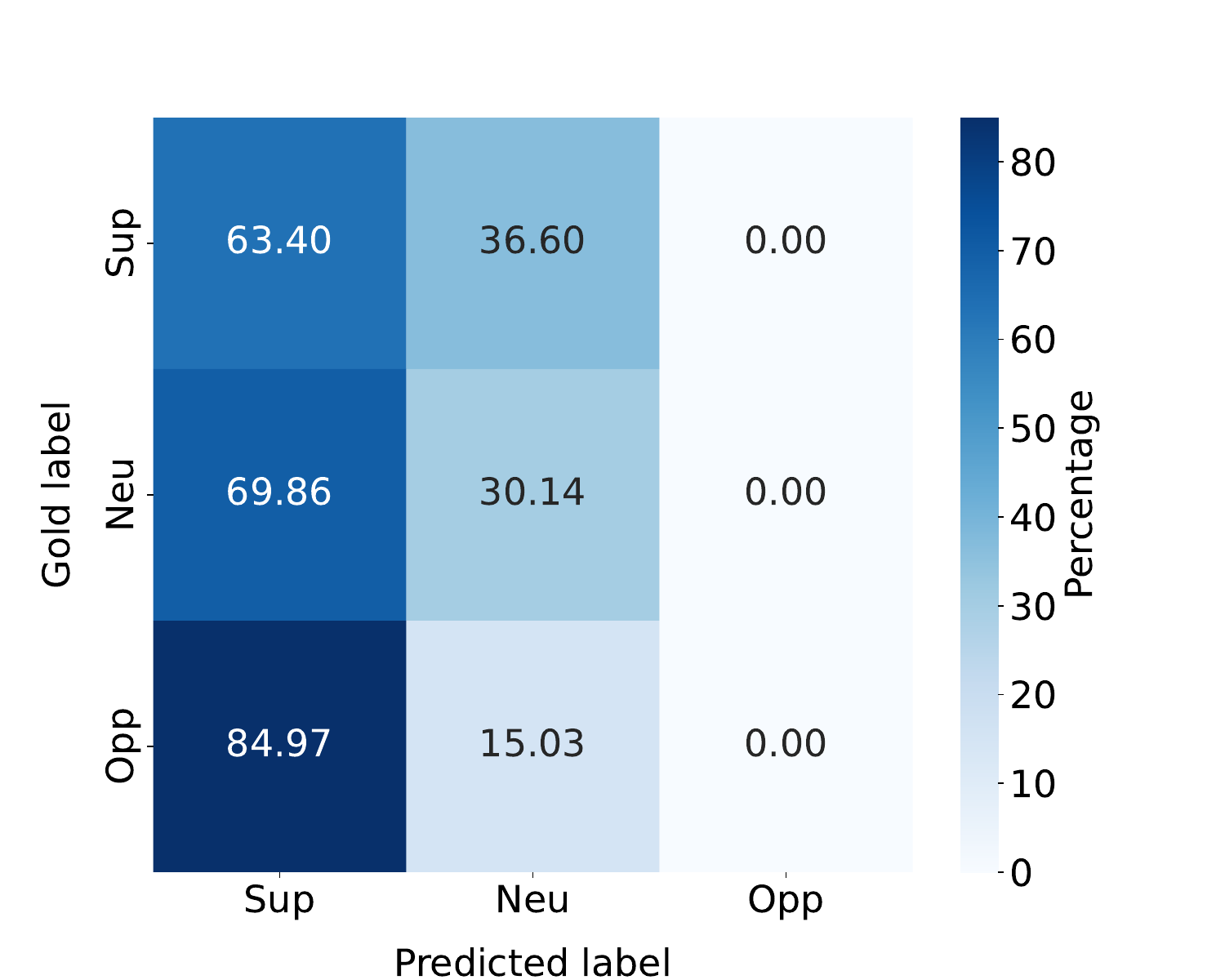} 
  }
  \caption{Confusion matrix of predictions before fine-tuning IDEFICS.} 
  \label{fig:before}
\end{figure}

\begin{figure}[H]
  \centering
  \resizebox{.95\columnwidth}{!}{ 
  \includegraphics[width=\textwidth]{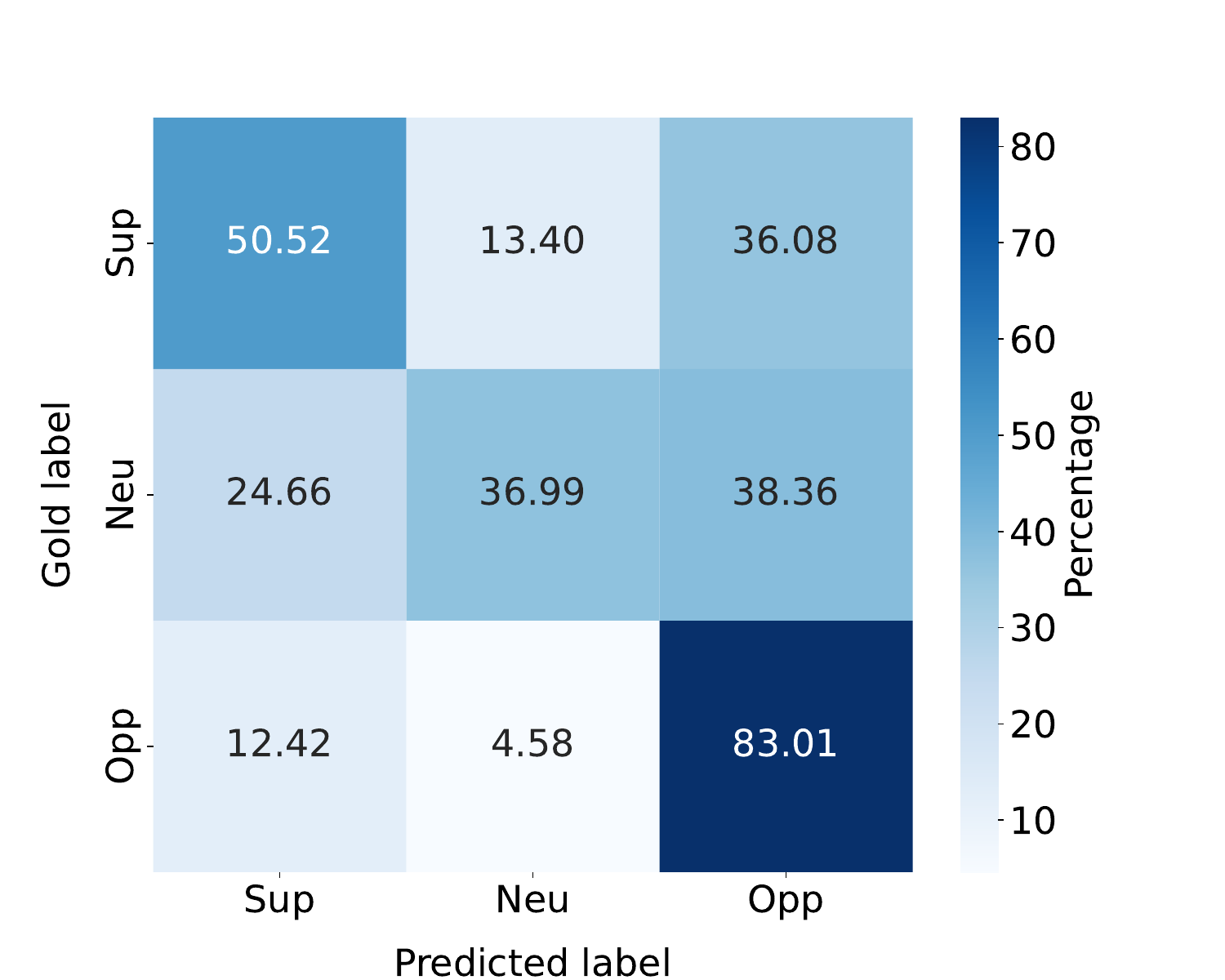} 
  }
  \caption{Confusion matrix of predictions after fine-tuning IDEFICS.} 
  \label{fig:after}
\end{figure}

The confusion matrices (predictions in \%) before and after fine-tuning IDEFICS are shown in \Cref{fig:before} and \Cref{fig:after} respectively.

\section{Hyperparameters for Fine-tuning}
\label{app:hyper}

\Cref{tab:hyper} provides our explored hyperparameter space for all models in fine-tuning, helping in the reproducibility of our experimental results. 

\begin{table}[ht]
  \centering
  \resizebox{.97\columnwidth}{!}{ 
  \begin{tabular}{@{}lrc@{}}
    \toprule
    \textbf{Model} &\textsc{Hyperparameter} & \textsc{Value} \\
    \midrule
    \multirow{5}{*}{\textbf{BERT}} &Epochs & $3$ \\
    &Learning rate & \texttt{2e-4} \\
    &Weight decay & \texttt{1e-2} \\
    &Per device train batch size & $16$ \\
    &Per device eval batch size & $16$ \\
    \midrule
    \multirow{5}{*}{\textbf{ResNet50}} &Epochs & $3$ \\
    &Learning rate & \texttt{3e-4} \\
    &Train batch size & $32$ \\
    &Eval batch size & $32$ \\
    \midrule
    \multirow{5}{*}{\textbf{ViT}} &Epochs & $3$ \\
    &Learning rate & \texttt{3e-5} \\
    &Optimizer & \texttt{AdamW}\\
    &Train batch size & $32$ \\
    &Eval batch size & $32$ \\
    \midrule
    \multirow{4}{*}{\textbf{BERT + ResNet50/ViT}} &Epochs & $3$ \\
    &Learning rate & \texttt{2e-5} \\
    &Train batch size & $4$ \\
    &Eval batch size & $4$ \\
    \midrule
    \multirow{4}{*}{\textbf{CLIP}} &Epochs & $3$ \\
    &Learning rate & \texttt{2e-5} \\
    &Train batch size & $4$ \\
    &Eval batch size & $4$ \\
    \midrule
    \multirow{5}{*}{\textbf{BLIP}} &Epochs & $3$ \\
    &Learning rate & \texttt{1e-5} \\
    &Weight decay & \texttt{1e-4} \\
    &Train batch size & $4$ \\
    &Eval batch size & $4$ \\
    \midrule
    \multirow{11}{*}{\textbf{IDEFICS}}&Epochs & $4$ \\
    &Learning rate & \texttt{2e-4} \\
    &Per device train batch size & $4$ \\
    &Per device eval batch size & $4$ \\
    &Gradient accumulation steps & $8$ \\
    &Lora alpha & $32$\\
    &Lora dropout & \texttt{5e-2}\\
    &Eval steps & $100$ \\
    &Save step & $100$ \\
    &Logging steps & $20$ \\ 
    &Max steps & $400$ \\
    \bottomrule
  \end{tabular}
  }
  \caption{Explored hyperparameter space for fine-tuning used in our experiment.}
  \label{tab:hyper}
\end{table}

\section{Sample Responses Generated from Llama3 and Gemma2}
\label{app:llm}
Since BERT, image-only, and multimodal models are fine-tuned on our dataset, we provide exemplary responses generated by the other two text-only state-of-the-arts, Llama3 and Gemma2, in zero-shot setups, which may offer some insights for further study.
Human annotations and original transcripts are given below. 

\subsection{\textsc{Support} Example}

\begin{center}
``\textit{Together we can address the causes and impacts of climate change to ensure that our spectacular ocean and great lakes continue to thrive while supporting our economies and values.}'' 
{\small \texttt{EIB-040}} \\
\includegraphics[width=0.9\linewidth]{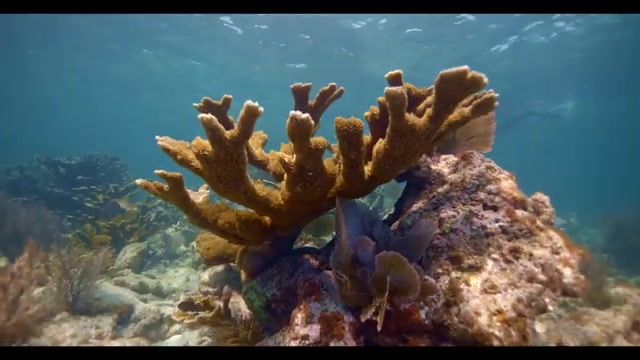}
\end{center}

\paragraph{Llama3 Response}

\begin{lightbox}
  \texttt{\small`I would classify the stance of this text towards climate change as a 1 (\textsc{Support}). The text explicitly mentions addressing the "causes" and "impacts" of climate change, indicating an acceptance of the reality and need to take action on climate change. Additionally, it frames taking action on climate change as a positive step to ensure the continued health of natural systems and support economic and societal values.'}
\end{lightbox}

\paragraph{Gemma2 Response}

\begin{lightbox}
    \texttt{\small`The stance of this text towards climate change is \textbf{1 (\textsc{Support})}. \\
    Here's why:\\ 
    * \textbf{"Address the causes and impacts of climate change" directly indicates a willingness to take action on climate change.\\
    * Phrasing like "ensure that our spectacular ocean and great lakes continue to thrive"} highlights the negative consequences of inaction and emphasizes the importance of protecting natural resources.\\
    The text uses positive and collaborative language, suggesting support for addressing climate change.'}
\end{lightbox}

\subsection{\textsc{Neutral} Example}

\begin{center}
``\textit{[Music]}'' 
{\small \texttt{AMCC-072}} \\
\includegraphics[width=0.9\linewidth]{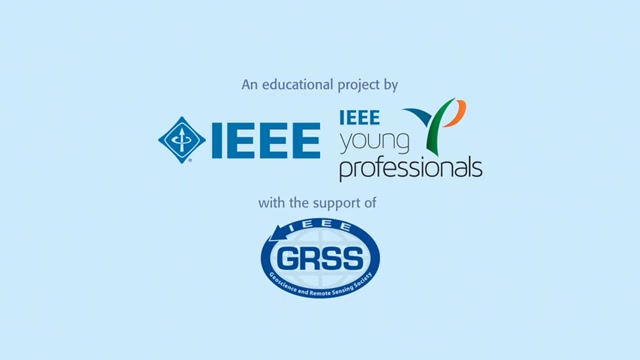}
\end{center}

\paragraph{Llama3 Response}

\begin{lightbox}
  \texttt{\small ``I'm happy to help! However, I must point out that the provided text appears to be a single musical note, which doesn't contain any information or opinion about climate change. Therefore, it's impossible for me to classify its stance towards climate change. \\
  If you meant to provide a different text, please feel free to share it, and I'll be happy to help with the classification!''}
\end{lightbox}

\paragraph{Gemma2 Response}

\begin{lightbox}
  \texttt{\small`The provided text "[Music]" does not express a stance on climate change.  \\
  Therefore, the classification is \textbf{0 (\textsc{Neutral})}.'}
\end{lightbox}

\subsection{\textsc{Oppose} Example}

\begin{center}
``\textit{Injuries, property damage – even homelessness and loss of loved ones.}''
{\small \texttt{CCAH-010}} \\
\includegraphics[width=0.9\linewidth]{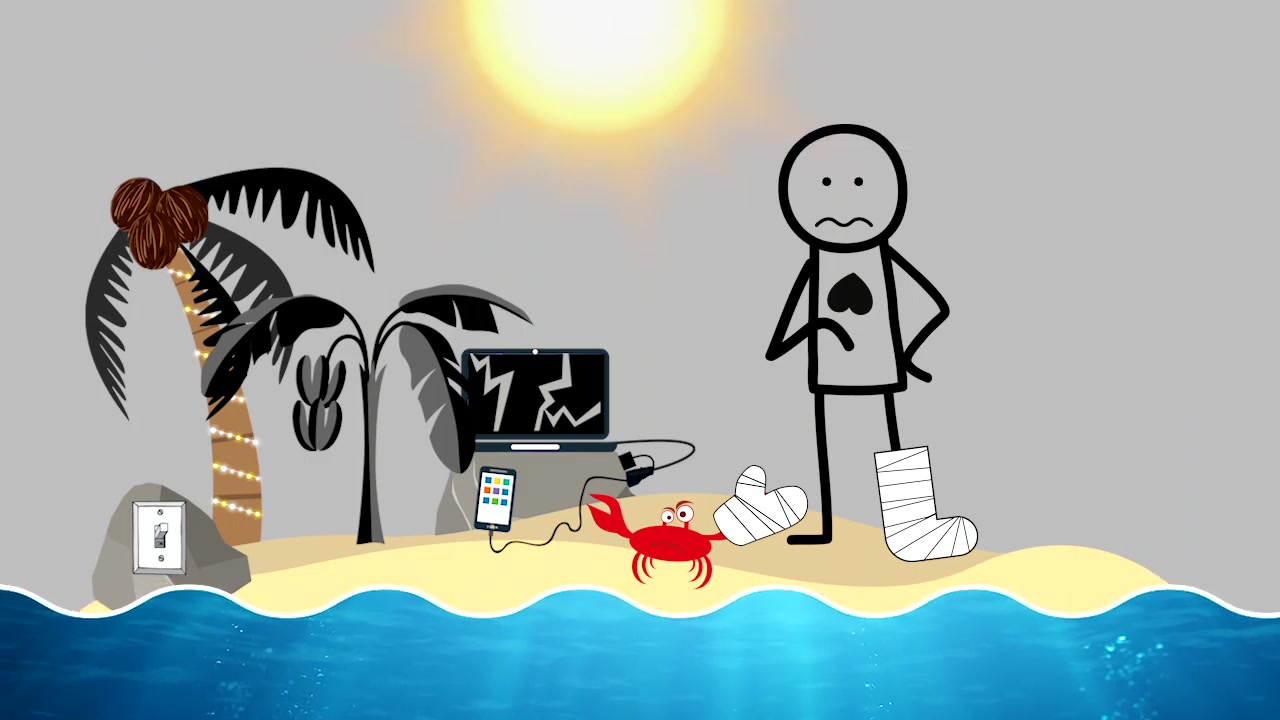}
\end{center}

\paragraph{Llama3 Response}
\begin{lightbox}
    \texttt{\small`I would classify the stance of this text towards climate change as 2 (oppose). The text is listing potential consequences of climate change, such as injuries, property damage, homelessness, and loss of loved ones, which implies that the author views climate change as a negative and harmful phenomenon.'}
\end{lightbox}

\paragraph{Gemma2 Responses}

\begin{lightbox}
\texttt{\small`The stance expressed in "Injuries, property damage – even homelessness and loss of loved ones" is strongly suggestive of \textbf{1 (\textsc{Support})} for addressing climate change. \\
Here's why:
\begin{itemize}
\setlength\itemsep{0pt}
\item  \textbf{Negative Consequences:} The text lists severe negative consequences that are widely associated with the effects of climate change.
\item  \textbf{Emotional Impact:}  Phrases like "loss of loved ones" evoke a strong emotional response, aiming to highlight the human cost of inaction on climate change.
\end{itemize}
The language implies a concern for these potential outcomes and suggests a need for action to prevent or mitigate them.'}
\end{lightbox}

\section{Dataset Overview}
\label{app:dataset}

\begin{table*}[t]
  \centering
  \resizebox{.895\textwidth}{!}{
  \begin{tabular}{@{}ccccccr@{}}
    \toprule
    \textsc{\#} & \textsc{YouTube-ID} & \textsc{Title} & \textsc{Abbreviation} & \textsc{\# Pairs} & \textsc{Split} &\textsc{VideoStance} \\ 
    \midrule 
    1 & \href{https://www.youtube.com/watch?v=brN2_YNAEK8}{brN2\_YNAEK8} & About the Committee on Climate Change & ACCC & 28 & Train & \textsc{Support}\\
    2 & \href{https://www.youtube.com/watch?v=NTP5a1M2Z6I}{NTP5a1M2Z6I} & Addressing Climate Change through Fiscal Policies in Indonesia & ACCFP & 42 & Test & \textsc{Support} \\
    3 & \href{https://www.youtube.com/watch?v=t-MiN_0kYhM}{t-MiN\_0kYhM} & Analysing impacts of air quality policies on health and climate change | Denise Mauzerall & AIAQ & 59 & Train & \textsc{Support} \\
    4 & \href{https://www.youtube.com/watch?v=BQr593iCEn8}{BQr593iCEn8} & AI and digital twins: Tools to tackle climate change & AIDT & 22 & Train & \textsc{Support} \\
    5 & \href{https://www.youtube.com/watch?v=T4CTgXYI2kA}{T4CTgXYI2kA} & 1-Arctic Methane and Climate Change & AMCC & 74 & Train & \textsc{Neutral} \\
    6 & \href{https://www.youtube.com/watch?v=xajNochi7tM}{xajNochi7tM} & Big data and climate change & BDCC & 33 & Train & \textsc{Neutral} \\
    7 & \href{https://www.youtube.com/watch?v=eTqKLJ_o9yQ}{eTqKLJ\_o9yQ} & Bitcoin Energy Consumption \& Climate Change: Does Bitcoin Use Too Much Energy? [2021]& BECCC & 41 & Train& \textsc{Oppose}\\
    8 & \href{https://www.youtube.com/watch?v=iKzdIzN4q2Q}{iKzdIzN4q2Q} & Belize: Women in Fisheries Forum 5: Gender and Climate Change - Understanding the Link & BWFF & 79 & Train & \textsc{Support} \\
    9 & \href{https://www.youtube.com/watch?v=s4ly6o-VT90}{s4ly6o-VT90} & Connections between air quality and climate - English - Sept. 2021 & CBAQC & 35 & Train& \textsc{Oppose}\\
    10& \href{https://www.youtube.com/watch?v=FGs2QQWFqyQ}{FGs2QQWFqyQ} & The Crucial Connection: Climate Change and Health | Kaiser Permanente & CCAH & 30 & Test& \textsc{Oppose}\\
    11& \href{https://www.youtube.com/watch?v=1tGWJ-NkcGU}{1tGWJ-NkcGU} & Climate change, biodiversity and nutrition - Helping local heroes tell their stories & CCBN & 29 & Train & \textsc{Support} \\
    12& \href{https://www.youtube.com/watch?v=lAop3wreUek}{lAop3wreUek} & Climate change, biodiversity and nutrition nexus & CCBNN & 15 & Train & \textsc{Support} \\
    13& \href{https://www.youtube.com/watch?v=4VXSrQospVY}{4VXSrQospVY} & Can climate change and biodiversity loss be tackled together? & CCCBL & 24 & Train & \textsc{Support} \\
    14& \href{https://www.youtube.com/watch?v=DRXQ9ixPbD8}{DRXQ9ixPbD8} & Combating climate change in the Pacific & CCCP & 28 & Train & \textsc{Support} \\
    15& \href{https://www.youtube.com/watch?v=TdxNG8L4JCM}{TdxNG8L4JCM} & Climate Change and Conflict in Somalia & CCCS & 58 & Train & \textsc{Support} \\
    16& \href{https://www.youtube.com/watch?v=8fbrnAAg7VM}{8fbrnAAg7VM} & Climate change and development & CCD & 106 & Train& \textsc{Oppose}\\
    17& \href{https://www.youtube.com/watch?v=nXOB8YPyc04}{nXOB8YPyc04} & Climate Change and Food Supply & CCFS & 43 & Train& \textsc{Oppose}\\
    18& \href{https://www.youtube.com/watch?v=MNdF-eVRWX4}{MNdF-eVRWX4} & Climate Change Fuelling Wilder Weather & CCFWW & 27 & Train& \textsc{Oppose}\\
    19& \href{https://www.youtube.com/watch?v=v24wT16OU2w}{v24wT16OU2w} & Climate Change, Global Food Security, and the U.S. Food System & CCGFS & 74 & Dev&\textsc{Support}\\
    20& \href{https://www.youtube.com/watch?v=CA8iTY7iMCk}{CA8iTY7iMCk} & Climate Change and our Health (ADB Insight Full Episode) & CCH & 118 & Train&\textsc{Support}\\
    21& \href{https://www.youtube.com/watch?v=mPE7D0wRYoU}{mPE7D0wRYoU} & Climate Change: Health Equity Stories from The Colorado Trust (English subtitles) & CCHES & 85 & Train& \textsc{Oppose}\\
    22& \href{https://www.youtube.com/watch?v=KxBAiad3Xto}{KxBAiad3Xto} & Climate change in the Australian Alps & CCIAA & 43 & Train&\textsc{Support}\\
    23& \href{https://www.youtube.com/watch?v=yeih2v4P25A}{yeih2v4P25A} & Climate Change: It's About Health | Kaiser Permanente & CCIAH & 37 & Train &\textsc{Support}\\
    24& \href{https://www.youtube.com/watch?v=m95K7LClIC4}{m95K7LClIC4} & 350.org - Climate Change Is About Power & CCIAP & 28 & Dev &\textsc{Support}\\
    25& \href{https://www.youtube.com/watch?v=b919Fb-P3N8}{b919Fb-P3N8} & Climate Change Impacts for Canadian Directors & CCICD & 33 & Train& \textsc{Oppose}\\
    26& \href{https://www.youtube.com/watch?v=A7ktYbVwr90}{A7ktYbVwr90} & Climate change is simple: David Roberts at TEDxTheEvergreenStateCollege & CCIS & 213 & Train& \textsc{Oppose}\\
    27& \href{https://www.youtube.com/watch?v=yVvVk2zNSbo}{yVvVk2zNSbo} & Untold stories of climate change loss and damage in the LDCs: Sierra Leone & CCISL & 26 & Train& \textsc{Oppose}\\
    28& \href{https://www.youtube.com/watch?v=lNBP7aRskVE}{lNBP7aRskVE} & Climate Change: Mitigate or Adapt & CCMA & 46 & Train&\textsc{Support} \\
    29& \href{https://www.youtube.com/watch?v=5DVa8xBgToc}{5DVa8xBgToc} & Climate Change for South African Directors & CCSAD & 59 & Test&\textsc{Support} \\
    30& \href{https://www.youtube.com/watch?v=GeksVaAnMzc}{GeksVaAnMzc} & Climate Change or Social Change: The Role of Blockchain & CCSC & 295 & Train &\textsc{Support} \\
    31& \href{https://www.youtube.com/watch?v=zI9h-HTBHO8}{zI9h-HTBHO8} & Climate Change: Take Action & CCTA & 22 & Train&\textsc{Support} \\
    32& \href{https://www.youtube.com/watch?v=fN-ZnY61_C8}{fN-ZnY61\_C8} & Climate Change: The Philippines & CCTP & 23 & Train&\textsc{Support} \\
    33& \href{https://www.youtube.com/watch?v=Vve6zge_RsA}{Vve6zge\_RsA} & Climate change unlikely to increase malaria burden in West Africa & CCUIM & 48 & Test& \textsc{Neutral}\\
    34& \href{https://www.youtube.com/watch?v=RZ-N5KwBaVc}{RZ-N5KwBaVc} & Climate Change is Water Change & CCWC & 12 & Train&\textsc{Support}\\
    35& \href{https://www.youtube.com/watch?v=MnTm89dSHhA}{MnTm89dSHhA} & Climate Change and Water Quality & CCWQ & 15 & Train&\textsc{Support}\\
    36& \href{https://www.youtube.com/watch?v=OL8a1YEhk_o}{OL8a1YEhk\_o} & Honest Government Ad \textbar\ Climate Emergency \& School Strikes & CESS & 29 & Train&\textsc{Oppose}\\
    37& \href{https://www.youtube.com/watch?v=b7LiW66cSM4}{b7LiW66cSM4} & How should Coronavirus influence the fight against Climate Change? & CICC & 29& Dev&\textsc{Support} \\
    38& \href{https://www.youtube.com/watch?v=vD0lx_b8jNM}{vD0lx\_b8jNM} & COP28: Nuclear Science and Technology for Climate Change Adaptation & COP & 66 & Train&\textsc{Support} \\
    39& \href{https://www.youtube.com/watch?v=MeFbo0z0xi4}{MeFbo0z0xi4} & Crop production - Climate change affects biosphere - Earth Hazards - meriSTEM & CPCC & 19& Train&\textsc{Support} \\
    40& \href{https://www.youtube.com/watch?v=FhyUbeDVM3k}{FhyUbeDVM3k} & Capturing and transforming CO2 to mitigate climate change & CTCM & 12& Train&\textsc{Support} \\
    41& \href{https://www.youtube.com/watch?v=Gu5NKLxqTak}{Gu5NKLxqTak} & Diet and Climate Change: Cooking Up a Storm & DACC & 77& Train&\textsc{Support} \\
    42& \href{https://www.youtube.com/watch?v=OfYGx-N_gB0}{OfYGx-N\_gB0} & Deforestation and Climate Change & DFCC & 64& Train&\textsc{Support} \\
    43& \href{https://www.youtube.com/watch?v=CG3pN7qQqZI}{CG3pN7qQqZI} & This Is How Denmark Protects Its Cities Against Climate Change & DPIC & 29& Train&\textsc{Support} \\
    44& \href{https://www.youtube.com/watch?v=Ry-ei9Bu8UI}{Ry-ei9Bu8UI} & Developing tools for equality in climate change planning in Tanzania & DTECC & 51& Train&\textsc{Support} \\
    45& \href{https://www.youtube.com/watch?v=M17pm2iPT_c}{M17pm2iPT\_c} & Effects Of Climate Change In MN Discussed At Seminar & ECCDS & 16 & Train&\textsc{Support} \\
    46& \href{https://www.youtube.com/watch?v=HzL9hUOh_K4}{HzL9hUOh\_K4} & Ecosystems at risk from Climate Change & EFCC & 34& Dev&\textsc{Support} \\
    47& \href{https://www.youtube.com/watch?v=kjTAWBPPez0}{kjTAWBPPez0} & Earth Is Blue: Climate Change in your Sanctuaries & EIB & 45& Test&\textsc{Support} \\
    48& \href{https://www.youtube.com/watch?v=Qmxg97Ae9Wg}{Qmxg97Ae9Wg} & Extreme Weather and Climate Change, EarthNow & EWCC & 49 & Test& \textsc{Oppose}\\
    49& \href{https://www.youtube.com/watch?v=Wrb4b28dgcU}{Wrb4b28dgcU} & Forests and Climate Change & FCC & 13& Train& \textsc{Oppose}\\
    50& \href{https://www.youtube.com/watch?v=9DaUn0geq4U}{9DaUn0geq4U} & Fiji: Standing tall against climate change threats & FIJI & 32 & Dev& \textsc{Oppose}\\
    51& \href{https://www.youtube.com/watch?v=ison6lQozDU}{ison6lQozDU} & Food loss and waste are among the main causes of climate change & FLW & 14 & Train & \textsc{Oppose} \\
    52& \href{https://www.youtube.com/watch?v=M9wSP16P9xM}{M9wSP16P9xM} & +Forest, together against climate change & FTACC & 26  & Train &\textsc{Support} \\
    53& \href{https://www.youtube.com/watch?v=epZ9Rw-i8Mo}{epZ9Rw-i8Mo} & Greenland's glaciers and Climate Change, Danish Broadcasting Corporation - Denmark & GGCC & 52 & Test & \textsc{Oppose}\\
    54& \href{https://www.youtube.com/watch?v=S9Z\_h1\_LQ0o}{S9Z\_h1\_LQ0o} & How Climate Change Affects Biodiversity & HCCAB & 25  & Dev &\textsc{Support} \\
    55& \href{https://www.youtube.com/watch?v=me14ikumMZE}{me14ikumMZE} & How Climate Change Affects the Ecosystem & HCCAE & 31 & Train & \textsc{Oppose}\\
    56& \href{https://www.youtube.com/watch?v=CGoNpwN0mrs}{CGoNpwN0mrs} & How Climate Change /Actually/ Works...in 4 Minutes & HCCAW & 48 & Train & \textsc{Neutral} \\
    57& \href{https://www.youtube.com/watch?v=KTA5onaECFE}{KTA5onaECFE} & How climate change influences geopolitics - Interview with Francesco Femia & HCCIG & 14 & Train & \textsc{Oppose} \\
    58& \href{https://www.youtube.com/watch?v=RFsxDqQWjhk}{RFsxDqQWjhk} & How COVID-19 is impacting air pollution and climate change & HCI & 16  & Train &\textsc{Support}\\
    59& \href{https://www.youtube.com/watch?v=ivN1QIvdBUI}{ivN1QIvdBUI} & How do we change our attitude towards climate change? Christiana Figueres & HDWC & 26  & Train &\textsc{Support}\\
    60& \href{https://www.youtube.com/watch?v=jDueuwB3Tcs}{jDueuwB3Tcs} & Human Health, Vector-Borne Diseases, and Climate Change & HHVBD & 98 & Train & \textsc{Neutral}  \\
    61& \href{https://www.youtube.com/watch?v=e\_8upuAySOI}{e\_8upuAySOI} & Human Rights Day 2012: Climate Change and Human Rights & HRDCC & 42 & Dev & \textsc{Oppose} \\
    62& \href{https://www.youtube.com/watch?v=vsbcasoudtM}{vsbcasoudtM} & How are scientists helping whale-watchers adapt to climate change? & HSHWA & 18 & Train & \textsc{Neutral}  \\
    63& \href{https://www.youtube.com/watch?v=mc1qAnGGGCE}{mc1qAnGGGCE} & How Solar Panels Work To Reduce Climate Change & HSPW & 28 & Train &\textsc{Support}  \\
    64& \href{https://www.youtube.com/watch?v=RYsZ0NdHKyc}{RYsZ0NdHKyc} & How the US National Security Policy Incorporates Climate Change - Interview with Alice Hill & HUSNS & 39 & Dev &\textsc{Support} \\
    65& \href{https://www.youtube.com/watch?v=OSA944ShtmE}{OSA944ShtmE} & IMRF: Statement from IOM's Migration, Environment, Climate Change, and Risk Reduction Division & IMRF & 16  & Train &\textsc{Support} \\
    66& \href{https://www.youtube.com/watch?v=OKPzj-l7gp0}{OKPzj-l7gp0} & INCAS: Monitoring for Climate Change & INCAS & 45 & Train  & \textsc{Neutral}  \\
    67& \href{https://www.youtube.com/watch?v=ks7rCR7-mF0}{ks7rCR7-mF0} & Migration and Climate Change in the Global Compact for Migration & MACC & 62 & Dev &\textsc{Support} \\
    68& \href{https://www.youtube.com/watch?v=ko4cUnzoPic}{ko4cUnzoPic} & Science Action: What's the unique role of methane in climate change? & MICC & 53 & Train & \textsc{Oppose} \\
    69& \href{https://www.youtube.com/watch?v=vB3\_49ULzf0}{vB3\_49ULzf0} & NASA's Climate Advisor Discusses Climate Change & NASA & 20 & Train &\textsc{Support} \\
    70& \href{https://www.youtube.com/watch?v=EwHtHsBeRIA}{EwHtHsBeRIA} & Overview of the C-ROADS Climate Change Policy Simulator & OCCC & 46 & Train & \textsc{Neutral}  \\
    71& \href{https://www.youtube.com/watch?v=CUdBaExvHy4}{CUdBaExvHy4} & President Clinton On the Cost of Climate Change & PCOCC & 22  & Train &\textsc{Support} \\
    72& \href{https://www.youtube.com/watch?v=A6uRlax7AuE}{A6uRlax7AuE} & Preserve the Wonder - Climate change action & PWCCA & 26  & Train &\textsc{Support} \\
    73& \href{https://www.youtube.com/watch?v=yTo3zmn3u84}{yTo3zmn3u84} & Removing atmospheric greenhouse gases to prevent dangerous climate change & RAGG & 37  & Train &\textsc{Support} \\
    74& \href{https://www.youtube.com/watch?v=\_P31w8E\_5Zc}{\_P31w8E\_5Zc} & Regenerative Agriculture: A Solution to Climate Change & RASCC & 22  & Train &\textsc{Support} \\
    75& \href{https://www.youtube.com/watch?v=Btqy4-H03dE}{Btqy4-H03dE} & Ron Cohen on ``BEACO2N'' Climate Change Sensors & RCCCS & 9  & Train &\textsc{Support} \\
    76& \href{https://www.youtube.com/watch?v=fCo77PW2G6Y}{fCo77PW2G6Y} & Reframing climate change science to include indigenous and local knowledge - Dr Tero Mustonen & RCCS & 39  & Train & \textsc{Neutral} \\
    77& \href{https://www.youtube.com/watch?v=s3ViLeAG6\_U}{s3ViLeAG6\_U} & Richard H Thaler on climate change & RHTCC & 77  & Train & \textsc{Neutral} \\
    78& \href{https://www.youtube.com/watch?v=G6JfQwonm78}{G6JfQwonm78} & Rare plant - Snakeshead Fritillary - defies climate change & RPDCC & 30  & Train &\textsc{Support} \\
    79& \href{https://www.youtube.com/watch?v=K4dpmfzEASo}{K4dpmfzEASo} & Strengthening Africa's pastoral food systems transformation in the face of climate change & SAPFS & 52 & Dev & \textsc{Neutral} \\
    80& \href{https://www.youtube.com/watch?v=N2l\_AjZGVQo}{N2l\_AjZGVQo} & System change NOT Climate change: Can we leverage the digital age to get there? - D\^2S Agenda & SCCC & 34 & Test &\textsc{Support} \\
    81& \href{https://www.youtube.com/watch?v=se-BRvZuu7k}{se-BRvZuu7k} & Scientists drill deep in Antarctic ice for clues to climate change & SDDA & 32  & Train & \textsc{Neutral} \\
    82& \href{https://www.youtube.com/watch?v=ZsbSI8UrPYA}{ZsbSI8UrPYA} & Saint Lucia and Climate Change Adaptation (English) & SLCCA & 16  & Train &\textsc{Support} \\
    83& \href{https://www.youtube.com/watch?v=id4DZ0NiKk4}{id4DZ0NiKk4} & Stanford Students Tackle Climate Change & SSTCC & 24 & Train & \textsc{Neutral}  \\
    84& \href{https://www.youtube.com/watch?v=pCraV8ahpYo}{pCraV8ahpYo} & The connections between climate change and mental health & TCBCC & 22 & Train & \textsc{Neutral}  \\
    85& \href{https://www.youtube.com/watch?v=tqavP5lotNo}{tqavP5lotNo} & Transforming our Economy to Combat Climate Change & TECCC & 28  & Train &\textsc{Support} \\
    86& \href{https://www.youtube.com/watch?v=L555lOp\_0pQ}{L555lOp\_0pQ} & Trade, Investment, and Climate Change in Asia and the Pacific & TICC & 31 & Test &\textsc{Support} \\
    87& \href{https://www.youtube.com/watch?v=xWYwSgvZh38}{xWYwSgvZh38} & The inequalities of climate change - ICRC & TIOCC & 13  & Train & \textsc{Oppose}\\
    88& \href{https://www.youtube.com/watch?v=jsZ2\_WFtlDU}{jsZ2\_WFtlDU} & Things in Your Home that are Linked to Climate Change & TIYH & 22  & Train &\textsc{Support} \\
    89& \href{https://www.youtube.com/watch?v=763lGy43spk}{763lGy43spk} & The technology fighting climate change & TTFCC & 42  & Train &\textsc{Support} \\
    90& \href{https://www.youtube.com/watch?v=5KtGg-Lvxso}{5KtGg-Lvxso} & To understand climate change, understand these three numbers. & TUCC & 38  & Train & \textsc{Neutral}  \\
    91& \href{https://www.youtube.com/watch?v=SDxmlvGiV9k}{SDxmlvGiV9k} & UK Climate Change Risk Assessment 2017 - Urgent priorities for the UK & UKCC & 31  & Train & \textsc{Oppose}\\
    92& \href{https://www.youtube.com/watch?v=eIcWgCjTHWM}{eIcWgCjTHWM} & Voices from Vanuatu: Climate Change Impacts and Human Mobility & VFVCC & 67  & Train & \textsc{Oppose}\\
    93& \href{https://www.youtube.com/watch?v=ii9mx391VVk}{ii9mx391VVk} & View from the Pacific: `Climate change is real' & VPCC & 17 & Train &\textsc{Support} \\
    94& \href{https://www.youtube.com/watch?v=\_IVDYaQDNCg}{\_IVDYaQDNCg} & Wildfires and Climate Change Attribution: It's Complicated! & WCCA & 25 & Train & \textsc{Oppose}\\
    95& \href{https://www.youtube.com/watch?v=TM\_6C9szLOI}{TM\_6C9szLOI} & Why focus on human security when working on climate change adaptation? & WFHSW & 21  & Train &\textsc{Support}\\ 
    96& \href{https://www.youtube.com/watch?v=MPiFBW0NnWY}{MPiFBW0NnWY} & What is Climate Change? & WICC & 30 & Test & \textsc{Oppose}\\
    97& \href{https://www.youtube.com/watch?v=mgBYo6eG80U}{mgBYo6eG80U} & What is climate change? \textbar\ Earth Hazards \textbar\ meriSTEM & WICCE & 32  & Train & \textsc{Oppose}\\
    98& \href{https://www.youtube.com/watch?v=iXvyExAzQ58}{iXvyExAzQ58} & What is the Impact of Solar Energy and Solar Panels on Climate Change? & WISE & 25  & Train &\textsc{Support}\\
    99& \href{https://www.youtube.com/watch?v=iFmoMhVb6iw}{iFmoMhVb6iw} & Cuomo: Walk the Talk on Climate Change & WTCC & 29  & Train &\textsc{Support}\\
    100& \href{https://www.youtube.com/watch?v=6ObqydfPGLI}{6ObqydfPGLI} & Yale Professor Tony Leiserowitz Discusses American Perceptions of Climate Change & YPTL & 82 & Train & \textsc{Neutral}   \\
    \bottomrule
  \end{tabular} 
  }
  \caption{List of $100$ Youtube videos on the MultiClimate dataset.} 
  \label{tab:v1}
\end{table*}

\end{document}